\documentclass[10pt,journal,compsoc]{IEEEtran}
\usepackage{amsmath,amsfonts}
\usepackage{algorithmic}
\usepackage{algorithm}
\usepackage{array}
\usepackage{hyperref}
\usepackage{xurl}
\usepackage[caption=false,font=normalsize,labelfont=sf,textfont=sf]{subfig}
\usepackage{textcomp}
\usepackage{stfloats}
\usepackage{url}
\usepackage{verbatim}
\usepackage{graphicx}
\usepackage{cite}
\usepackage{multirow}
\hypersetup{
     colorlinks=true,
     linkcolor=black,
     filecolor=black,
     citecolor = black,      
     urlcolor=black,
     }

\begin{document}

\title{Test Against High-Dimensional Uncertainties: Accelerated Evaluation of Autonomous Vehicles with Deep Importance Sampling}

% \title{Test Against High-Dimensional Uncertainties: Certifiable Accelerated Evaluation of Autonomous Vehicles with Deep Importance Sampling}
%\title{Certifiable Evaluation of Autonomous Vehicles using Deep Importance Sampling (Deep IS)}

%
%\author{Authors,~\IEEEmembership{Carnegie Mellon University}
        % <-this % stops a space
%\thanks{This paper was produced by the IEEE Publication Technology Group. They are in Piscataway, NJ.}% <-this % stops a space
%\thanks{Manuscript received April 19, 2021; revised August 16, 2021.}
%}

\author{Mansur Arief, Zhepeng Cen, Zhenyuan Liu, Zhiyuan Huang, Henry Lam, Bo Li, Ding Zhao% <-this % stops a space
\IEEEcompsocitemizethanks{\IEEEcompsocthanksitem Mansur Arief, Zhepeng Cen, and Ding Zhao are with the Department
of Mechanical Engineering, Carnegie Mellon University, Pittsburgh,
PA, USA.
% \IEEEcompsocthanksitem Chejian Xu was with the Computer Science Department, Zhejiang University, Hangzhou, Zhejiang, China. e-mail: \texttt{xuchejian@zju.edu.cn}.\protect\\
\IEEEcompsocthanksitem Zhenyuan Liu and Henry Lam are with the Department of Industrial Engineering and Operations Research, Columbia University, New York City, NY, USA.
\IEEEcompsocthanksitem Zhiyuan Huang is with Tongji University, China.
\IEEEcompsocthanksitem Bo Li is with the Department of Computer Science, University of Illinois Urbana Champaign (UIUC), Urbana Champaign, IL, USA.
}% <-this % stops a space
%\thanks{This is a preprint version.}

}



\maketitle

\begin{abstract}
Evaluating the performance of autonomous vehicles (AV) and their complex subsystems to high precision under naturalistic circumstances remains a challenge, especially when the failure or dangerous cases are rare. Rarity does not only require an enormous sample size for a naive method to achieve high confidence estimation, but it also causes dangerous underestimation of the true failure rate and it is extremely hard to detect. Meanwhile, the state-of-the-art approach that comes with a correctness guarantee can only compute an upper bound for the failure rate under certain conditions, which could limit its practical uses. In this work, we present Deep Importance Sampling (Deep IS) framework that utilizes a deep neural network to obtain an efficient IS that is on-par with the state-of-the-art, capable of reducing the required sample size 43 times smaller than the naive sampling method to achieve  10\% relative error and while producing an estimate that is much less conservative. Our high-dimensional experiment estimating the misclassification rate of one of the state-of-the-art traffic sign classifiers further reveals that this efficiency still holds true even when the target is very small, achieving over 600 times efficiency boost. This highlights the potential of Deep IS in providing a precise estimate even against high-dimensional uncertainties. %Furthermore, we show that by cognizantly using the risk upper-bound and the risk threshold as a sampling termination criterion, we can achieve an additional 5.6 times acceleration boost compared to only using Deep IS estimation for the unbiased target, which significantly improve the efficiency while avoiding the risk of underestimating the target failure rate.
\end{abstract}


\section{Introduction}

\IEEEPARstart{I}{nnovations} driven by the synthesis of artificial intelligence (AI) algorithms and robotics designs in the development of intelligent cyber-physical systems have shown performance with tremendous potentials. Autonomous vehicles (AVs), for instance, are now driving alongside human drivers on public roads \cite{av_deployment}. AVs are equipped with various kinds of sensors, actuators, and complex AI algorithms, all working together to handle various tasks including perception, decision making, and controls. AI enables efficient processing of large streams of high-dimensional input data, supplying its predictions to downstream tasks, such as the decision-making modules to actualize proper driving maneuvers. While the resulting performance has been promising and appealing, it is necessary to ensure that these complex AI-driven technologies always perform reliably and ultra safely when deployed  \cite{national2017automated, kiran2021deep}. The question of safety is therefore becoming the forefront issue to tackle in order for the public to reap the full potentials of AVs such as reduced congestion and improved road safety \cite{CAV_UK, NHTSA_safety}.

One important element of safety assurance is  a rigorous evaluation framework prior to deployment. The rigor in evaluation methods plays a huge role in ensuring that not only the evaluated system passed the minimum safety bar (e.g. safer than human driver), but also the precision and uncertainties of the evaluation results are properly taken into consideration. Addressing evaluation uncertainty is particularly important for safety-critical applications since the failure rate is often small (e.g. 1.26 fatal cases per 100 million miles of driving \cite{nhtsa_2021}), but comes with serious consequences \cite{tesla_crash}, thus should not just be ignored.

In this regard, traditional evaluation methods are deemed insufficient for AVs. Formal verification, which analyzes and verifies the design using rigorous mathematical analysis, is often limited to a simpler representation of the AVs and their surroundings to maintain tractability \cite{rizaldi2016formally}. \cite{pegasus, jacobo2019development} often omit common and `boring' scenarios which could ignore part of dangerous cases and reduce the test scope \cite{riedmaier2020survey}. Function-based approaches, which are widely used for ADAS, are impractical for AV as a whole system as it is nearly impossible to enumerate all the  required functionality of AVs in every plausible situation\cite{riedmaier2020survey}. 

Meanwhile, the more recent evaluation approaches are also subject to their own limitations. Shadow testing approach \cite{shadow_mode}, in which the AV processes sensory data, make decisions, and then compares them with the human driver's decisions, could only analyzes the similarity between AI decisions and and human decisions. As a result, one could not evaluate the the real consequences of the AI decisions, hence lacking informativeness. Staged introduction (e.g. geofencing \cite{geofencing}), that tests and limits the operational design domains of AVs, only suits testing up to SAE Level 4 autonomy \cite{sae_level, riedmaier2020survey}. Finally, naturalistic field operational test, i.e. letting the AVs roam on public roads and encountering  naturalistic human drivers under various driving conditions, is well-known to be very expensive and inefficient, especially if one were aiming to achieve meaningful statistical significance \cite{Kalra2014}. This problem still holds true even in computer-simulated environments. The sample size required to simulate blows up very quickly as the probability of encountering dangerous cases becomes rarer  \cite{zhao2018accelerated}.

%could not produce reliable performance estimates with sufficient precision for AI-empowered systems due to their vast input and parameter space. 

Our work is part of studies concerning the development of efficient evaluation methods for AI-empowered cyber-physical systems. Works in this area can be categorized into two major categories: falsification and probabilistic evaluation \cite{corso2020survey}. Falsification mainly deals with situations in which worst-case scenarios are unknown and should be avoided at all costs. Traditionally, falsification runs a thorough examination of every potential failure case through exhaustive search or directly preventing them by design to occur through formal methods \cite{thurston1991formal, kobuna2016validation, zhang2017software}. Adversarial learning techniques have recently gained attention to scale this up to high-dimensional complex models but require some gradient information \cite{1247923} of the model. Successes of such an approach have been reported across many domain applications, including traffic sign identification \cite{213847, 459121}, and LiDAR point cloud object detection \cite{238272}.
Probabilistic evaluation, on the other hand, views the problem differently. Rather than aiming at finding all the failure cases, the goal is to estimate the overall probability of encountering failure cases in the real world. This can be achieved by considering the disturbance as a random variable, i.e. generated by some naturalistic distribution that represents the stochasticity of the environment.  Monte Carlo methods and its accelerated version using importance sampling have been proposed for this task to compute safety metrics estimators using a large number of samples \cite{zhao2015accelerated, huang2018versatile}. The main limitation of these methods is that they can only handle a low-dimensional problem and requires some level of information of the underlying systems. An extension that could be used for complex or black-box systems is proposed in \cite{arief2021certifiable}, but can only be used to estimate an upper-bound for the target risk metric.

%Practically, the latter direction provides a more practical measure to quantify and assess the risk of deployment. This is especially so because, in many applications with complex autonomous systems, it is often impossible or too costly to guarantees 100\% safety. For safety-critical applications, however,  the tolerated failure rate for systems in this area is extremely low. This also means that the test results are often extremely imbalanced, in which failure cases are very rarely encountered during tests. This rarity often poses a practical challenge since it requires test methods to have an accelerated feature to make the cost for testing affordable. Test methods lacking this notion of acceleration could not keep up with rare faulty cases and could render rigorous testing prohibitively expensive. 

In this work, we adopt the risk upper-bound estimator from \cite{arief2021certifiable} to compute a less biased estimator to solve the evaluation task. This is particularly important for evaluating and comparing the safety of AV prototypes whose performances are often really close to one another. In this situation, comparing an upper bound of the risk is far less meaningful compared to directly comparing their true risks. We call the framework Deep Importance Sampling (Deep IS). We show using an evaluation of AV agent and a traffic sign classifier that Deep IS outputs a risk estimate that is very close to the true target and can achieve an acceleration rate of 650 times. This means the sampling size required by naive Monte Carlo methods is hundreds of times more than the sample size required by Deep IS. We also benchmark Deep IS with the risk upper bound estimator (that we call Robust Deep IS in this paper) as well as its iterative extension that we name Iter. Robust Deep IS. These risk upper-bound estimators provide robust estimates by its conservative nature, providing additional safety measure against the many unknowns of these AI-empowered systems deployment in real-world.

Our contribution is threefold, outlined as follows:
\begin{itemize}
    \item we propose a practical framework to design efficient importance sampler for large-dimensional problem, capable of evaluating AV traffic sign classifier
    \item we extend the use of the certifiable risk-upper bound estimator by using iterative sampling method and relaxed outer approximation, significantly reducing its degree of conservativeness, and
    \item we achieve significant efficiency improvement that highlight the effectiveness and practicality of our framework on high-dimensional rare-event problem and highlight the tradeoff between efficiency and correctness of various IS-based estimators.% (autonomous vehicle safety evaluation, 60 dimensions) and  traffic sign classifier for autonomous driving agents (4,096 dimensions). 
\end{itemize}
This work is an extension to an earlier version submitted to a conference \cite{arief2022certifiable} that focuses only on evaluating AV traffic sign classifier. We add an additional benchmark, the Iterative Robust Deep IS (Alg. \ref{alg:it_robust_deep_is}), as an improvement to the risk upper-bound estimator proposed in \cite{arief2020deep}.  In addition, we conduct more experiments: AV driving controller evaluation under cut-in-tailgate scenario and an ablation study of the traffic sign classifier evaluation, varying the rarity parameter to better study the efficiency trend of Deep IS method against rarer failure cases.

The rest of this paper is organized as follows. In Section \ref{sec:problem_formulation}, we present the problem setting and formulation and provide a short overview of related work in the literature. In Section \ref{sec:framework}, we describe our framework and present algorithms that serve as benchmarks. Section \ref{sec:exp} describes our experiments. %where autonomous driving evaluation example is presented in Subsection \ref{sec:exp_1} and traffic sign classifier evaluation is presented in Subsection \ref{sec:exp_2}.
Finally, we discuss our findings in Section \ref{sec:discussion} and conclude in Section \ref{sec:conclusion}.

\section{Problem formulation}
\label{sec:problem_formulation}

\begin{figure}[t]
    \centering
    \includegraphics[width=\linewidth]{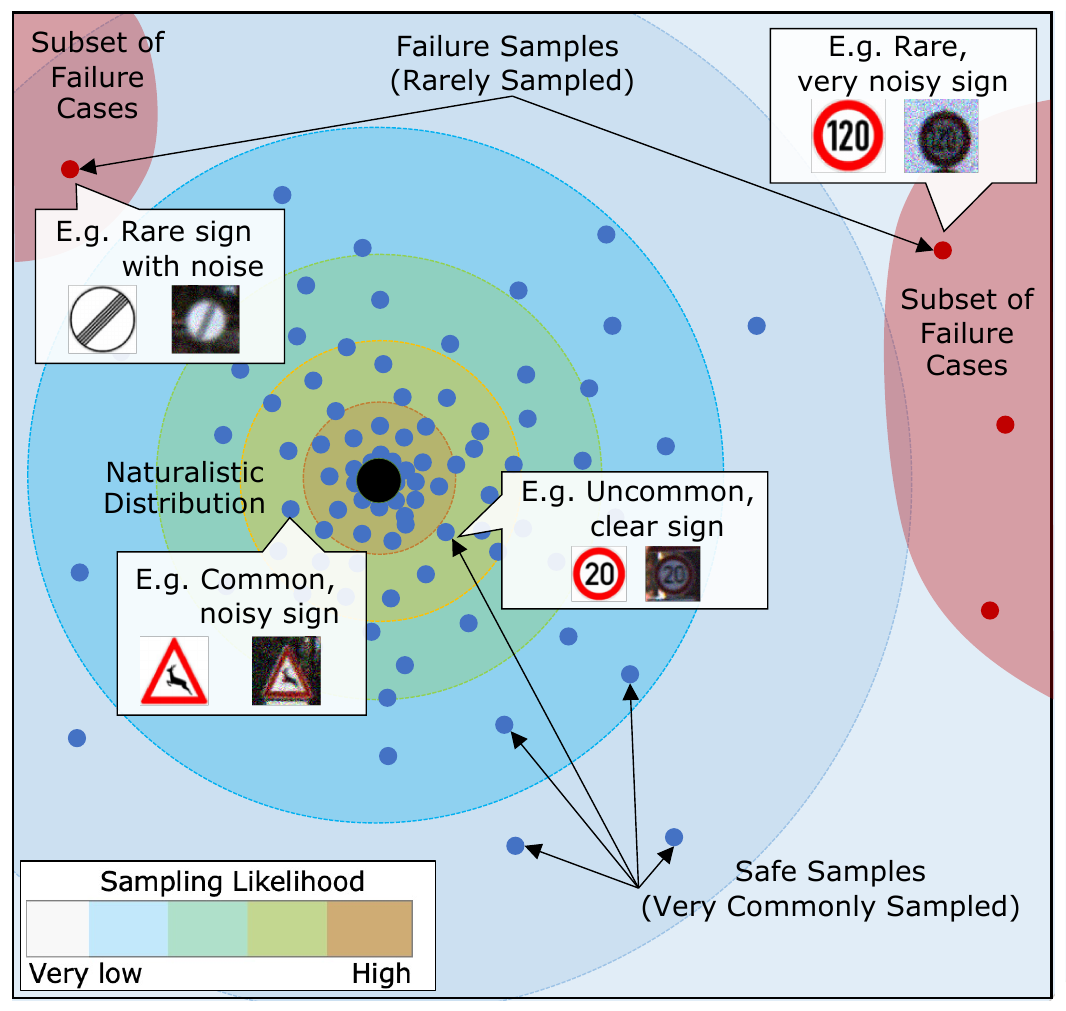}
    \caption{Illustration of rare-event problem in evaluating the performance of traffic sign classifier for self-driving cars. In this case, the rare dangerous subsets contain signs that are likely misclassified by a trained deep learning classifiers (e.g. traffic signs with low representation in the training set or common signs subjected to very noisy perturbations).}
    \label{fig:setting}
\end{figure}

%%%%%make larger
\begin{figure*}[t]
    \centering
    \includegraphics[width=\linewidth]{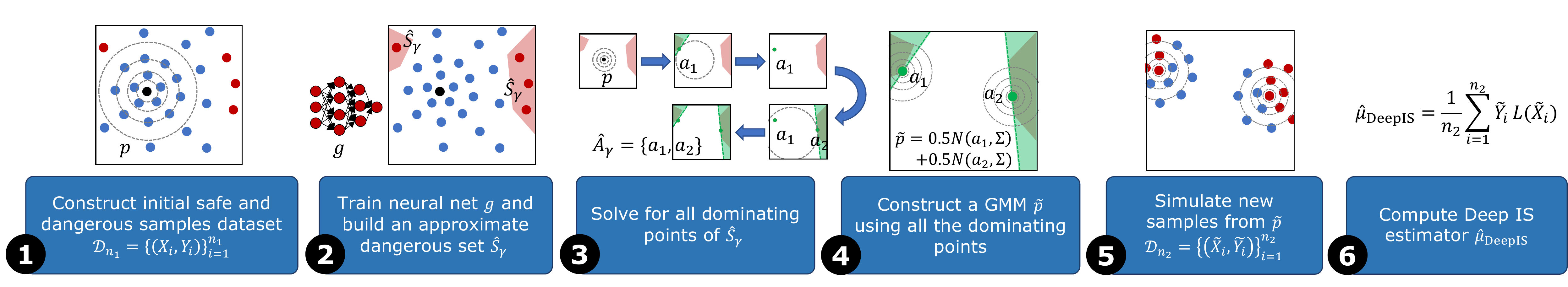}
    \caption{Illustration of Deep IS Framework.}
    \label{fig:framework}
\end{figure*}

We adopt the formulation of the safety of AVs from the safety testing literature as how likely it encounters dangerous scenarios when operating under naturalistic circumstances \cite{zhao2015accelerated, huang2018versatile, feng2021intelligent}. The notion of naturalistic here means that the environment behaviour, represented here by a multivariate random variable $X \in \mathbb R^d$, follows a naturalistic randomness, modeled by some probability distribution $p$.  The dangerous events, say $\mathcal S \subset \mathbb R^d$, is a subset of all realizations which lead the AV to trigger dangerous behaviour to itself or the surrounding environment, which include pedestrian, other road users, or physical infrastructure. For example, for AV perception subsystem, such as traffic sign classifier, the set may contain rare and noisy traffic sign images (see illustration in Fig. \ref{fig:setting}. The true topology and nature of $\mathcal S$ can be really complex, but it is often helpful to parameterize it with some parameter $\gamma$, (hence written as $\mathcal S_\gamma$), considering it could change as AV technological evolves or public behavior shifts towards AV navigating on public roads \cite{elliott2019recent}. %For instance, as illustrated in Fig. \ref{fig:setting}, the dangerous cases for a traffic sign classifier may consist rare traffic signs with vision perturbations.

With this setting, the above safety notion of the AV can be formulated as
\begin{equation}
\mu = \mathbb P(X \in \mathcal S_\gamma) = \mathbb E_{X \sim p}[1_{X \in \mathcal S_\gamma}],
\end{equation}
where $1_\xi$ is the indicator function with value 1 if the statement $\xi$ is true, and 0 otherwise and the expectation is taken with respect to naturalistic distribution $p$. Thus, $\mu$ represents the probability that the AV encounters dangerous situations under naturalistic randomness of its operational design domains (ODD). This notion of safety is deemed helpful because it allows comparing the risk of deploying AVs with that of human drivers (e.g. one such statistic is 1.26 fatal cases per 100 millions miles of driving \cite{nhtsa_2021}) and ranking the alternative designs during the design process.

\subsection{Sample Size Requirement for AV Testing}

A safety testing problem aims to estimate $\mu$ using an estimator $\hat \mu_n$,  computed using $n$ samples, with sufficient accuracy and precision. More precisely, the sample size requirement $n$ is related to the target confidence level of our final estimator $\hat \mu_n$. Suppose that we aim to achieve $\epsilon$ relative error w.r.t. $\mu$ with high confidence and use $n$ samples $Z_1, \cdots, Z_n$ where $Z_i = 1_{X_i \in \mathcal S_\gamma}$. Since the target $\mu$ can be tiny, the error between $\hat \mu_n$ and $\mu$ should be measured relative to the magnitude of $\mu$. This is particularly important since the the estimation is only meaningful if the error is small enough relative to $\mu$. With this regard, the sample size requirement $n$ ensures
\begin{equation}
P(|\hat\mu_n-\mu|>\epsilon \mu)\leq\delta,
\label{goal}
\end{equation} 
where $\delta$ is our target confidence level (e.g., $\delta=5\%$) and $0<\epsilon<1$.
A straightforward application of Markov inequality reveals that using crude sampling approach (i.e. Monte Carlo), the number of samples $n$ required to achieve (\ref{goal}) must satisfy 
\begin{equation}
n \geq \frac{\mathrm{Var}(Z_i)}{\mu^2 \delta \epsilon^2},
\label{eq:nsample}
\end{equation} 
which can be prohibitively large for sufficiently safe systems (i.e. when $\mu \approx 0$). To see this more clearly, note that $Z_i$ has a Bernoulli distribution with success probability $\mu$, hence $\text{Var}(Z_i) = \mu(1-\mu)$. Thus, $n$ blows up quickly as $\mu \to 0$. This problem is well-known as rare-event challenge in safety evaluation task \cite{zhao2018accelerated}.

\subsection{Accelerated Evaluation using Importance Sampling}

Importance Sampling (IS) has been successfully applied to improve sampling efficiency for rare-event problems \cite{bucklew2013introduction}. The main idea is to tilt the probability measure toward more dangerous events. This is realized by generating more dangerous samples $X$ from another distribution $\tilde p$ (instead of $p$), and outputs the sample average weighted by the likelihood ratio $L$
\begin{equation}
    L=dp/d\tilde p.
\end{equation} 
The critical part of this approach is to carefully select the proposal distribution $\tilde p$ so that by using this alternate estimates, one can achieve a small relative error with a much smaller sample size. 

One known algorithm to realize this is to use a mixture model that account for the most likely scenarios that cover all the local modes of dangerous sets. These most likely scenarios are called dominating point in the rare-event literature \cite{sadowsky1990large,dieker2005asymptotically}. This idea has been adopted by earlier work in this direction under various settings. In \cite{7534875}, the authors derived an efficient IS distribution for the evaluation of simple lane-change AV scenarios, an improvement to the earlier work \cite{zhao2015accelerated} that uses heuristic approaches. In \cite{zhao2018accelerated}, the authors extended this approach to a dynamic system and evaluated car-following scenarios. In earlier works, a unimodal parametric class is used so that cross-entropy formulation is sufficient to solve for the IS optimal change of measure efficiently. The relaxation is proposed using a simple parametric class to include piecewise models in \cite{7989024, 8116682} and Gaussian Mixture Models in \cite{huang2018versatile} with theoretical guarantees provided in \cite{huang2018rare} adopting the dominating point idea along with iterative cutting plane methods. \cite{arief2021certifiable} applies the method for black-box systems, computing an upper-bound for the failure probability of the system with an efficiency guarantee. The limitation of existing studies is that they all focus on developing methods for relatively low-dimensional decision-making algorithms usually based on control theories and decision trees which have been lagged behind by the increasingly complex AI algorithms, e.g. basic longitudinal and lateral control \cite{ulsoy2012automotive, mcityreport}. Larger systems are addressed in \cite{feng2021intelligent} at the cost of level of rigor. Here, we synthesize these approaches and propose Deep IS framework that achieves high efficiency boost for relatively high-dimensional problem. %Our framework is derived based on the same idea, but aiming at generalizing the framework to settings beyond the theoretical exposition in \cite{arief2020deep} that assumes certain topology for the set $\mathcal S_\gamma$.

\section{Deep IS Framework and Description of Benchmarks}
\label{sec:framework}

The key ingredient of our framework is the versatile use of deep learning classifier to approximate the dangerous set $\mathcal S_\gamma$. We illustrate the framework in Fig. \ref{fig:framework}. We assume to have collected some dataset of size $n_1$ that can be used for learning the shape of dangerous events (Step 1). For simplicity, here we assume that $X \sim p$, i.e. the environment behaviour surrounding AV follows some naturalistic distribution $p$. The learning procedure trains a deep neural net $g$ for classification/set learning task, extracting prior information from dataset $\mathcal D_{n_1} = \{(X_i, Y_i)\}_{i = 1}^{n_1}$ comprised of normal cases ($Y_i = 0$) and dangerous cases ($Y_i = 1$), where $Y_i = 1_{X_i \in \mathcal S_\gamma}$ (Step 2).  
Assuming that $p=N(\lambda, \Sigma)$, we then solve for the set of dominating points $\hat A_\gamma$ (Step 3) and construct our proposal mixture model 
\begin{equation}
\tilde p = \frac{1}{|\hat A_\gamma|} \sum_{a \in \hat A_\gamma}N(a, \Sigma),
\end{equation}
where $N(a, \Sigma)$ is a Gaussian distribution with mean $a$ and covariance $\Sigma$ (Step 4). Finally, we perform final sampling, collecting $n_2$ samples and their label $\mathcal D_{n_2} = \{(\tilde X_i, \tilde Y_i)\}_{i=1}^{n_2}$ (Step 5) and compute Deep IS estimator
\begin{equation}
    \hat \mu_{\text{DeepIS}} = \frac{1}{n_2} \sum_{i=1}^{n_2} L(\tilde X_i) \tilde Y_i,
\end{equation}
where 
\begin{equation}
L(\tilde X_i) =  \frac{\phi (\tilde  X_i;\lambda,\Sigma)}{ \frac{1}{|\hat A_\gamma|}\sum_{a\in \hat A_\gamma} \phi (\tilde  X_i;a,\Sigma)},
\end{equation}
$\tilde Y_i = 1_{\tilde X_i \in \mathcal S}$, and $\phi(\cdot, \lambda, \Sigma)$ is the density of Gaussian distribution with mean $\lambda$ and covariance $\Sigma$ (Step 6).

The algorithm for Deep IS framework is summarized in Alg. \ref{alg:deep_is}. We present a more detailed description here. In Stage 1, we train a deep neural network classifier, say $g$, using all $n_1$ samples $\mathcal D_{n_1} = \{(X_i,Y_i)\}_{i=1}^{n_1}$, and obtain an approximate rare-event region
\begin{equation}
\hat{\mathcal S}_\gamma =\{x: g(x) \geq 0\}.
\end{equation}
Leveraging on the strength of deep learning to learn complex decision boundary, we assume that proper training gives us $\hat{\mathcal S}_\gamma \approx \mathcal S_\gamma$. %With this, we can run mixture IS to estimate $P(X\in \hat{\mathcal  S}_\gamma)$ in Stage 2. 

Since efficient IS proposal necessitates a mixture model that accounts for all dominating points of $\hat{\mathcal S}_\gamma$, we need an efficient programming to solve them in a finite time. To achieve this, we adopt  efficient mixed integer program (MIP) formulation for ReLU neueral network \cite{tjeng2017evaluating, huang2018designing}. When $g$ is a ReLU-activated neural net,  $\hat{\mathcal S}_\gamma$ is a union of polytopes and its boundary is piecewise linear. In particular, suppose we use $L$-layer deep neural network with ReLU activation functions and the $i$-th layer outputs of ReLU denoted as 
\begin{equation}
s_i = \max\{W_i^T s_{i-1} + b_i, 0\}.
\end{equation}
Then, we can replace the constraint $g(x) \geq 0$ in (\ref{eq:opt_ite}) with
\begin{align}
    s_L & \geq 0, &\\
    s_i &\leq W_i^T s_{i-1} + b_i + M(1-z_i), & i=1,\cdots, L \\
    s_i &\geq W_i^T s_{i-1}+b_i, &i=1, \cdots, L \\
    s_i &\leq Mz_i, & i = 1, \cdots, L\\
    s_i &\geq 0, & i=1, \cdots, L \\
    z_i & \in \{0, 1\}^{m_i}, & i=1, \cdots, L\\
    s_0 &= x,
\end{align}
where $W_i$'s and $b_i$'s are the weight and bias parameters of the neural network, $m_i$ is the number of nodes in $i$-th layer, and $M$ is some practical upper-bound value. The problem size with this formulation grows linearly to the network size (the number of integer variables is equal to the number of neurons and the number of constraints is roughly 4 times the number of neurons). 

Under this condition, the procedure to find all its dominating points boils down to  a sequential programming that iteratively locates the next dominating point by minimizing the rate function of the naturalistic distribution $p$ over the remaining portion of $\hat{\mathcal S}_\gamma$ not covered by the local region of any previously found points $x^\dagger$. In Alg. \ref{alg:deep_is}, we use the case of Gaussian $p = N(\lambda, \Sigma)$, with corresponding rate function 
\begin{equation}
I(x) = (x-\lambda)^T \Sigma^{-1} (x-\lambda)
\end{equation}
as objective. More general formulation is available in the literature, for instance formulation in \cite{arief2021certifiable}.

%Description of the framework: 
%\begin{itemize}
%    \item deep learning classifier for set learning
%    \item optimization problem for dominating points
%    \item IS estimator with GMM sampler (Deep IS)
%    \item Upper bound estimator with outer approximation prediction (Deep IS UB)
%    \item Iterative method (iterative Deep IS)
%    \item Mode switching: Deep IS $\longleftrightarrow$ Deep IS UB for efficient termination criterion
%\end{itemize}

\begin{algorithm}[H]
\caption{Deep IS}
\begin{algorithmic}[1]
\STATE {\textsc{Set learning with $n_1$ samples:}}
\STATE \hspace{0.1cm} Train classifier with + prediction  $\hat{\mathcal  S}_\gamma = \{ x: g(x) \geq 0 \}$ 
%\STATE \hspace{0.1cm} Replace $\kappa$ with $\hat\kappa=\max\{\kappa\in\mathbb R: (\overline{\mathcal  S}_\gamma^{\kappa})^c\subset\mathcal H(T_0)\}$

\vspace{0.3cm}
\STATE {\textsc{Dominating set w.r.t. $p = N(\lambda, \Sigma)$ and $\hat{\mathcal S}_\gamma$:}}
\STATE \hspace{0.1cm}  Initiate $\hat A_\gamma \leftarrow \emptyset, \mathcal X \leftarrow \hat{\mathcal S}_{\gamma}$

\STATE \hspace{0.1cm}  \textsc{While} $ \mathcal X \neq  \emptyset$:

\STATE \hspace{0.3cm} Solve the optimization problem \begin{align} \label{eq:opt_ite}
    x^* =\arg \min_{x} &\ \  (x-\lambda)^T \Sigma^{-1}(x-\lambda) \\
\text{s.t.}\ \ \  & g(x) \geq 0,\ \  \nonumber \\ &(x^\dagger-\lambda)^T\Sigma^{-1}(x-x^\dagger) <0, \ \mbox{$\forall x^\dagger \in\hat A_\gamma$} \nonumber
\end{align}

\STATE \hspace{0.3cm} Update $\hat A_\gamma \leftarrow\hat A_\gamma \cup \{x^*\}$
\STATE \hspace{0.3cm} Update $\mathcal X \leftarrow \mathcal X \cap_{x^\dagger \in \hat A_\gamma} (x^\dagger-\lambda)^T\Sigma^{-1}(x-x^\dagger) <0$
\STATE \hspace{0.1cm}  \textsc{End While}

\vspace{0.3cm}
\STATE {\textsc{Computing Estimator with $n_2$ samples:}}

\STATE \hspace{0.1cm}  Sample $\tilde X_1, \cdots ,\tilde X_{n_2}$ from   $\tilde p = \frac{1}{|\hat A_\gamma|} \sum_{a\in\hat A_\gamma} N(a,\Sigma) $

\STATE \hspace{0.1cm}  Compute $L(\tilde X_i)= \frac{\phi (\tilde  X_i;\lambda,\Sigma)}{ \frac{1}{|\hat A_\gamma|}\sum_{a\in \hat A_\gamma} \phi (\tilde  X_i;a,\Sigma)}$

\STATE \hspace{0.1cm}  Compute
$\hat\mu_{\text{DeepIS}}= \frac{1}{n_2} \sum_{i =1}^{n_2} L(\tilde  X_i) 1_{\tilde  X_i \in \mathcal{S}_{\gamma}}$

\end{algorithmic}
\label{alg:deep_is}
\end{algorithm}

\subsection{Benchmark \#1: Robust Deep IS}
We benchmark the proposed Deep IS framework with two other accelerated evaluation methods. First is the Deep Probabilistic Accelerated Evaluation approach first proposed in \cite{arief2020deep}, which we called Robust Deep IS here. The  `robustness' of this approach stems from its nature of estimating an upper-bound of the safety risk, instead of the true target, providing some level of hedging against unmeasurable risk. Robust Deep IS is very similar to our approach, but uses the prediction of the trained neural network $g$ with well-tuned decision threshold $\hat \kappa$ to construct an outer approximation $\overline{\mathcal S}_\gamma \supseteq \mathcal S_\gamma$ instead of $\hat {\mathcal S}_\gamma$. 
The tuning of $\hat \kappa$ ensures that the positive (+1) prediction region $\{x: g(x) \geq \hat \kappa \}$ contains the orthogonally monotone hull for all the Stage 1 samples that are dangerous
\begin{equation}
\mathcal H(T_0)= \mathcal H(\{X_i:Y_i=0\}).
\end{equation}
Here, we call a set $\mathcal S_\gamma \subset\mathbb R_+^d$ orthogonally monotone if for any two points $x,x'\in\mathbb R_+^d$, we have $x\leq x'$ (where the inequality is defined coordinate-wise) and $x\in\mathcal S_\gamma$ implies $x'\in\mathcal S_\gamma$ too. Furthermore, for a set of points $D=\{X_1,\ldots,X_n\}\subset\mathbb R_+^d$, 
we define the orthogonally monotone hull of $D$ (with respect to the origin) as $\mathcal H(D)=\cup_i\mathcal R(X_i)$, where $\mathcal R(X_i)$ is the rectangle that contains both $X_i$ and the origin as two of its corners. To put it simply, the orthogonally monotone hull consists of the union of all the rectangles each wrapping each point $X_i$ and the origin $0$.

With this, if $\mathcal S_\gamma$ is orthogonally monotone, then $\mathcal H(T_0)\subset\mathcal S_\gamma^c$, or equivalently, $\mathcal H(T_0)^c\supset\mathcal S_\gamma$, i.e., $\mathcal H(T_0)^c$ is an outer approximation of the rare-event set $\mathcal S_\gamma$. Note that $\mathcal H(T_0)^c$ is the smallest region that attains zero false negative rate, because any smaller region could exclude a point that has label +1  with positive probability. 

The use of outer-approximation guarantees the theoretical efficiency of Robust Deep IS estimator when the dangerous set $\mathcal S_\gamma$ is orthogonally monotone. The robustness of this estimator is due to its interesting use-case to estimate an upper bound $\hat{\mu}_{\text{RobustDeepIS}}$ that is guaranteed to avoid underestimating the risk \cite{arief2020deep}, with a property that $\mathbb E\left[\hat \mu_{\text{RobustDeepIS}}\right] \geq \mu$ . The tuning method for $\hat \kappa$ provides a way to minimize the conservativeness of the estimator, hence to some extent, can be viewed as a minimax estimator for the risk (i.e. minimizing the conservativeness of the maximum estimate for the probability of encountering a dangerous situations in the real world). The algorithm for Robust Deep IS is presented in Alg. \ref{alg:robust_deep_is}.

\begin{algorithm}
\caption{Robust Deep IS \cite{arief2020deep} \label{alg:robust_deep_is}}
\begin{algorithmic}[1]
\STATE {\textsc{ Outer-approximation with $n_1$ samples:}}
\STATE \hspace{0.1cm} Train classifier with + prediction  $\overline{\mathcal  S}_\gamma^\kappa = \{ x: g(x) \geq \kappa \}$ 

\STATE \hspace{0.1cm} Replace $\kappa$ with $\hat\kappa=\max\{\kappa\in\mathbb R: (\overline{\mathcal  S}_\gamma^{\kappa})^c\subset\mathcal H(T_0)\}$

\vspace{0.3cm}
\STATE {\textsc{Dominating set w.r.t. $p = N(\lambda, \Sigma)$ and $\overline{\mathcal S}_\gamma^{\hat \kappa}$:}}
\STATE \hspace{0.1cm}  Initiate $\hat A_\gamma \leftarrow \emptyset, \mathcal X \leftarrow \overline{\mathcal S}_{\gamma}^{\hat \kappa}$

\STATE \hspace{0.1cm}  \textsc{While} $ \mathcal X \neq  \emptyset$:

\STATE \hspace{0.3cm} Solve the optimization problem \begin{align*}
    x^* =\arg \min_{x} &\ \  (x-\lambda)^T \Sigma^{-1}(x-\lambda) \\
\text{s.t.}\ \ \  & g(x) \geq \hat \kappa,\ \  \nonumber \\ &(x^\dagger-\lambda)^T\Sigma^{-1}(x-x^\dagger) <0, \ \mbox{$\forall x^\dagger \in\hat A_\gamma$} \nonumber
\end{align*}

\STATE \hspace{0.3cm} Update $\hat A_\gamma \leftarrow\hat A_\gamma \cup \{x^*\}$
\STATE \hspace{0.3cm} Update $\mathcal X \leftarrow \mathcal X \cap_{x^\dagger \in \hat A_\gamma} (x^\dagger-\lambda)^T\Sigma^{-1}(x-x^\dagger) <0$
\STATE \hspace{0.1cm}  \textsc{End While}
\vspace{0.3cm}
\STATE {\textsc{ Computing Estimator with $n_2$ samples:}}

\STATE \hspace{0.1cm}  Sample $\tilde X_1, \cdots ,\tilde X_{n_2}$ from   $\tilde p = \frac{1}{|\hat A_\gamma|} \sum_{a\in\hat A_\gamma} N(a,\Sigma) $

\STATE \hspace{0.1cm}  Compute $L(\tilde X_i)= \frac{\phi (\tilde  X_i;\lambda,\Sigma)}{ \frac{1}{|\hat A_\gamma|}\sum_{a\in \hat A_\gamma} \phi (\tilde  X_i;a,\Sigma)}$

\STATE \hspace{0.1cm}  Output 
$\hat\mu_{\text{RobustDeepIS}}= \frac{1}{n_2} \sum_{i =1}^{n_2} L(\tilde  X_i) 1_{g(\tilde  X_i) \geq \hat \kappa}$

\end{algorithmic}
\label{alg1}
\end{algorithm}

\subsection{Benchmark \#2: Iterative Robust Deep IS}
One limitation of Robust Deep IS is that the exceedingly fast growth of the estimator conservativeness with respect to the dimension $d$. This is mainly due to the growing challenge of providing Stage 1 samples that are sufficiently exploring the boundary of $\mathcal S_\gamma$. To alleviate this issue, we propose an extension to Robust Deep IS algorithm. Rather than providing a static set of Stage 1 samples, we allow the algorithm to iteratively collect Stage 1 samples in $k$ batches. For $j$-th batch, we collect $n_{1, j} = n_1/k$ samples, learn a preliminary outer-approximation, find its dominating points, and collect another batch of samples surrounding these dominating points, and repeat. This simple extension appears to improve the quality of the outer approximation and significantly reduce the conservativeness of the resulting estimator. We called the estimator with this iterative procedure as Iterative Deep IS estimator ($\hat \mu_{\text{IterRobustDeepIS}}$). The corresponding algorithm is provided in Alg. \ref{alg:it_robust_deep_is}. %We present the numerical result of Deep IS, Robust Deep IS, and Iterative Robust Deep IS in Section \ref{sec:exp}.

\begin{algorithm}
\caption{Iterative Robust Deep IS \label{alg:it_robust_deep_is}}
\begin{algorithmic}[1]
\STATE {Prepare the samples $\mathcal D_{n_{1,1}} = \{(X_i, Y_i)\}_{i=1}^{n_{1,1}}$}
\STATE {\textsc{ Repeat for} $j=1, \cdots, k-1$:}
\STATE \hspace{0.1cm} Training a classifier $g$ using dataset $\mathcal D_{n_{1,1}}$ 

\vspace{0.1cm}
\STATE \hspace{0.25cm}{\textsc{Preliminary Dominating set:}}
\STATE \hspace{0.25cm}  Initiate $\hat A_\gamma \leftarrow \emptyset, \mathcal X \leftarrow \hat{\mathcal S}_{\gamma}$

\STATE \hspace{0.25cm}  \textsc{While} $ \mathcal X \neq  \emptyset$:

\STATE \hspace{0.5cm} Solve the optimization problem \begin{align*}
    x^* =\arg \min_{x} &\ \  (x-\lambda)^T \Sigma^{-1}(x-\lambda) \nonumber \\
\text{s.t.}\ \ \  & g(x) \geq 0,\ \  \nonumber \\ &(x^\dagger-\lambda)^T\Sigma^{-1}(x-x^\dagger) <0, \ \mbox{$\forall x^\dagger \in\hat A_\gamma$} \nonumber
\end{align*}
\STATE \hspace{0.5cm} Update $\hat A_\gamma \leftarrow\hat A_\gamma \cup \{x^*\}$
\STATE \hspace{0.5cm} Update $\mathcal X \leftarrow \mathcal X \cap_{x^\dagger \in \hat A_\gamma} (x^\dagger-\lambda)^T\Sigma^{-1}(x-x^\dagger) <0$
\STATE \hspace{0.25cm}  \textsc{End While}
\STATE \hspace{0.25cm}  Sample $\hat X_{1}, \cdots , \hat X_{n_{1,j+1}}$ from   $\frac{1}{|\hat A_\gamma|} \sum_{a\in\hat A_\gamma} N(a,\Sigma) $

\STATE \hspace{0.25cm}  Construct next training set $$\mathcal D_{n_1, j+1} = \mathcal D_{n_1, j} \cup \{(\hat X_i, 1_{\hat X_i \in \mathcal S_\gamma})\}_{i=1}^{n_{1,j+1}}$$

\STATE \hspace{0.1cm}{\textsc{ End For}}

\STATE \hspace{0.1cm} Train using  $\mathcal D_{n_1, k}$ with + region  $\overline{\mathcal  S}_\gamma^\kappa = \{ x: g(x) \geq \kappa \}$

\STATE \hspace{0.1cm} Replace $\kappa$ with $\hat\kappa=\max\{\kappa\in\mathbb R: (\overline{\mathcal  S}_\gamma^{\kappa})^c\subset\mathcal H(T_0)\}$

\vspace{0.3cm}
\STATE {\textsc{Final Dominating set w.r.t. $p = N(\lambda, \Sigma)$ and $\overline{\mathcal S}_\gamma^{\hat \kappa}$:}}
\STATE \hspace{0.1cm}  Initiate $\hat A_\gamma \leftarrow \emptyset, \mathcal X \leftarrow \overline{\mathcal S}_{\gamma}^{\hat \kappa}$

\STATE \hspace{0.1cm}  \textsc{While} $ \mathcal X \neq  \emptyset$:

\STATE \hspace{0.3cm} Solve the optimization problem \begin{align*}
    x^* =\arg \min_{x} &\ \  (x-\lambda)^T \Sigma^{-1}(x-\lambda) \\
\text{s.t.}\ \ \  & g(x) \geq \hat \kappa,\ \  \nonumber \\ &(x^\dagger-\lambda)^T\Sigma^{-1}(x-x^\dagger) <0, \ \mbox{$\forall x^\dagger \in\hat A_\gamma$} \nonumber
\end{align*}

\STATE \hspace{0.3cm} Update $\hat A_\gamma \leftarrow\hat A_\gamma \cup \{x^*\}$
\STATE \hspace{0.3cm} Update $\mathcal X \leftarrow \mathcal X \cap_{x^\dagger \in \hat A_\gamma} (x^\dagger-\lambda)^T\Sigma^{-1}(x-x^\dagger) <0$
\STATE \hspace{0.1cm}  \textsc{End While}

\vspace{0.3cm}
\STATE {\textsc{ Computing Estimator with $n_2$ samples:}}

\STATE \hspace{0.1cm}  Sample $\tilde X_1, \cdots ,\tilde X_{n_2}$ from   $\tilde p = \frac{1}{|\hat A_\gamma|} \sum_{a\in\hat A_\gamma} N(a,\Sigma) $

\STATE \hspace{0.1cm}  Compute $L(\tilde X_i)= \frac{\phi (\tilde  X_i;\lambda,\Sigma)}{ \frac{1}{|\hat A_\gamma|}\sum_{a\in \hat A_\gamma} \phi (\tilde  X_i;a,\Sigma)}$

\STATE \hspace{0.1cm}  Output 
$\hat\mu_{\text{IterRobustDeepIS}}= \frac{1}{n_2} \sum_{i =1}^{n_2} L(\tilde  X_i) 1_{g(\tilde  X_i) \geq \hat \kappa}$

\end{algorithmic}
\label{alg1}
\end{algorithm}

\subsection{Constructing $\mathcal D_{n_1}$ with Dangerous Scenario Synthesis}

In this subsection, we provide more details on generative model and learning-based approaches to create safety-critical scenarios, which are extremely helpful for construction set-learning datasets $\mathcal D_{n_1}$ as well as to  reveal the limitation of current algorithms, since the decision-making and perception tasks usually involve complicated environments that are not easily identified. Then we can generate new near-miss trajectories by interpolating in the latent space. Those generated trajectories can be used to evaluate the algorithms.
 
Generative models estimate the joint distribution of data and other attributes. They can learn the underlying distribution of the data and map the distribution to a low-dimensional latent space. To generate new data that also belongs to the same distribution, we can sample from the latent space and then transfer it to the data space. Based on different training procedures, generative models can be divided into three categories. The first kinds of model maximize the data likelihood directly with the change of variable rule. The second kinds optimize the evidence lower bound of the likelihood, which is an approximation. The third kinds do not optimize the likelihood but use the game theory to find an equilibrium point between a generative model and a discriminator. A good review is provided in \cite{ding2022survey}.

Another learning-based approach that has been widely used is Adaptive Stress Testing (AST) \cite{koren2018adaptive}. They search the traffic corner cases by Monte Carlo Tree Search and deep reinforcement learning method. The model controls a pedestrian to cross a road so their output is the control command at each step. Then, the pedestrian and vehicle interact in the simulator. The simulator will output a risk value that is corresponding to the action, and the target of their model is to maximize this risk value. Another work that shares a very similar structure is proposed by \cite{ding2020learning}. Different from AST, this method designs a Bayesian network structure (a kind of generative model) to represent a simple traffic scenario. The scenario includes the initial position and velocity of a cyclist. The target of the Bayesian network is to minimize the distance between the cyclist and the vehicle. This work explores the structural generative model for safety-critical scenario generation.

\section{Numerical experiments}
\label{sec:exp}

We provide two examples: a low-dimensional AV safety evaluation on cut-in-tailgate scenario and a high-dimensional problem on AV traffic sign classifier.

\subsection{AV Driving Controller Safety Evaluation}
\label{sec:exp_1}

%add info if one vehicle, both methods would work fine

\begin{figure}
    \centering
    \includegraphics[width=\linewidth]{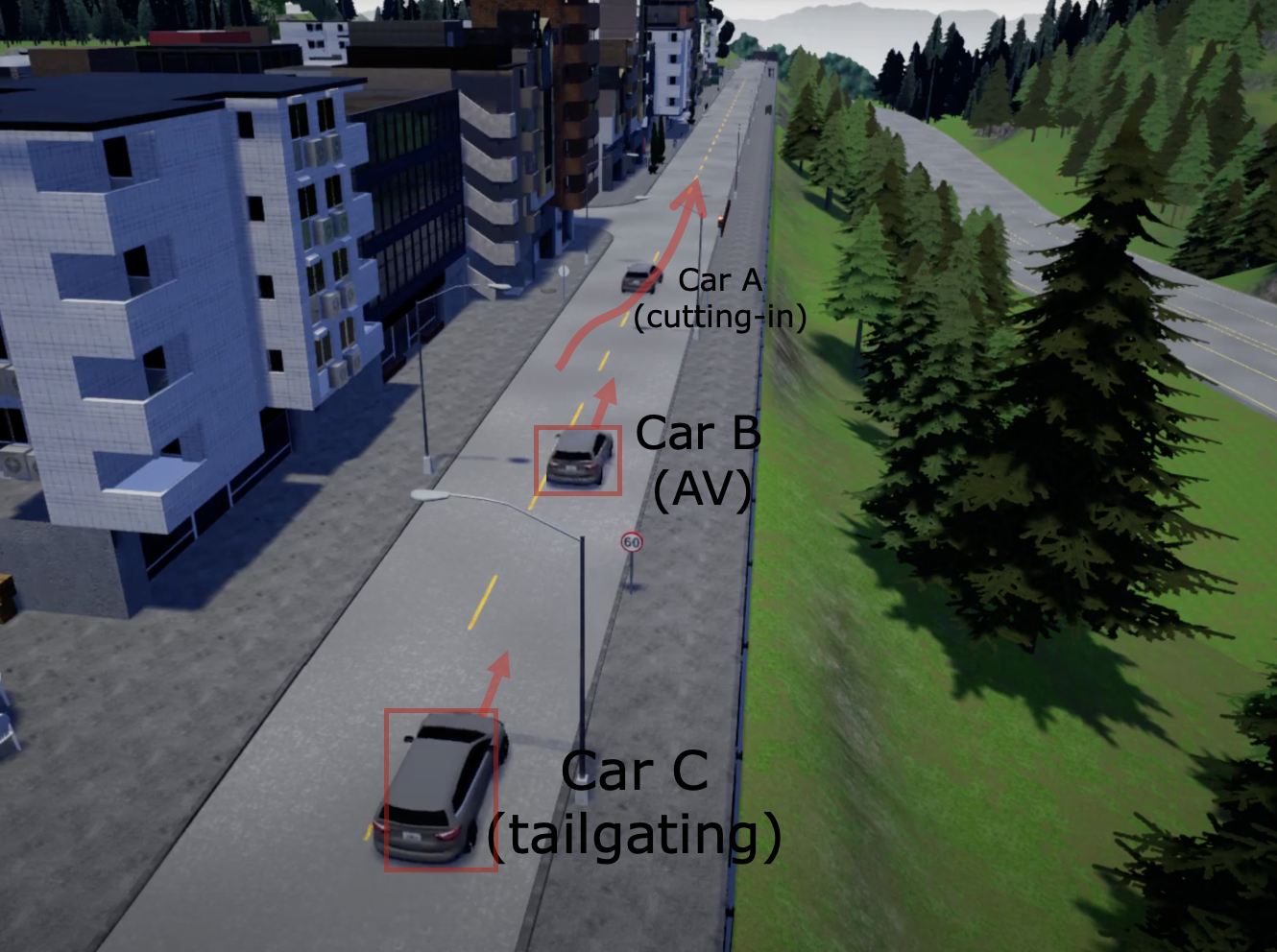}
    \caption{Cut-in-tailgate driving scenario illustration in Carla: AV facing a cut-in maneuver in the front and a tailgate by another vehicle in the rear}
    \label{fig:av_illustration}
\end{figure}

In this example, we first provide a low-dimensional problem that deals with evaluating the crash probability of followed cut-in scenarios involving a surrounding vehicle (Car A) making a lane-change in front of an autonomous vehicle (car B), followed by another surrounding vehicle (Car C). This scenario is common in highway driving, especially involving platooning \cite{aramrattana2017simulation} and often causes rear-end collisions \cite{rear_end}. We call the scenario cut-in-tailgate driving scenario for simplicity, with an illustration we simulated using Carla \cite{dosovitskiy2017carla} in Fig. \ref{fig:av_illustration}. In this regard, the AV uses a simple controller that maintains a safety distance while ensuring smooth ride and maximum efficiency \cite{treiber2000congested}.

The AV considers the latitude and longitudinal distances and relative velocities between itself and all the $k$ cars surrounding it when making decision. The relative distances and relative velocities (both longitudinal and lateral) of the surrounding vehicles are the observations in each time step $t =1, \cdots, T$ are treated as random variables $X \in \mathbb R^{4 \cdot k \cdot T}$, which follow some naturalistic distribution $p$. The steering angle and gas pedal in each time steps are the actions of the AVs. Here we consider driving scenarios when the AV is cruising originally at constant speed in the right-most lane, with another car trying to make a cut-in pass while at the same time another car is following the AV.

We are interested in estimating the probability that the AV hits either the front or rear car within $T$ timesteps . We denote the set of such crashes as $\mathcal S$ and express our goal as estimating $\mu = P(\cup_{i=1}^T X_i \in \mathcal S)$. As the AV becomes more and more capable in handling such situation, $\mu \to 0$, and thus the setting of rare-event kicks in. In the experiment, we set $k=2$, $T=15$, and each time step lasts for 0.6 seconds, which is about the same as average human reaction time in driving \cite{hugemann2002driver}. Hence, each simulation replication runs for 9 seconds. 

\begin{figure*}
    \centering
    \includegraphics[width=\linewidth]{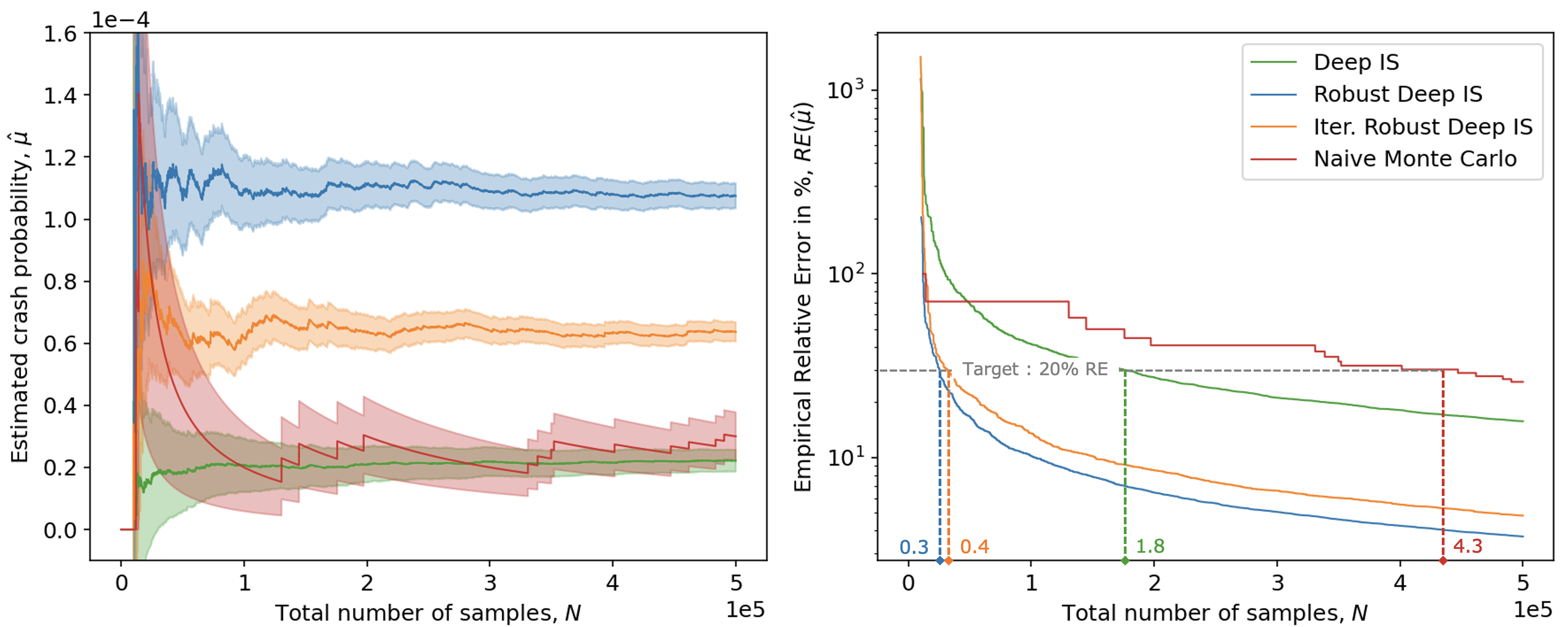}
    \caption{Estimation of probability of crashes for AV under cut-in-tailgate scenario}
    \label{fig:av_exp}
\end{figure*}

 We summarize our result in Fig. \ref{fig:av_exp}, showing the estimated crash probability from Naive Monte Carlo (NMC), Deep IS, Robust Deep IS, and Iter. Robust Deep IS. We also show the relative error of the estimates vs the number of samples used. In addition, we report the sample size required by each algorithm to achieve 20\% RE in Table \ref{tab1}. From this, we calculate the acceleration rate for Deep IS method
 \begin{equation}
     Acc\_Rate = \frac{n_{\text{NMC}}}{n_{\text{DeepIS}}},
 \end{equation}
 i.e. the ratio between NMC sample size $n_{\text{NMC}}$ to each of the Deep IS methods $n_{\text{DeepIS}}$.  We also report in Table \ref{tab1} the degree of conservativeness 
 \begin{equation}
     Deg\_Conservativeness = \frac{\hat \mu_\text{DeepIS}}{\mu},
 \end{equation}
 which is the ratio between Deep IS estimators to the target value (computed using NMC with much larger samples, i.e. 1 million samples). The former statistic will be used to compare the efficiency attained by each method (higher $Acc\_Rate$ means more efficient method). The latter is purported to compare the magnitude of each estimator to the true target ($Deg\_Conservativeness > 1$ means the estimated value $>$ target, and $Deg\_Conservativeness <$ 1 means the estimated value $<$ target).

We compare the estimator for $\mu$ from Naive Monte Carlo sampling, Deep IS, Robust Deep IS, and Iter. Robust Deep IS in Fig. \ref{fig:av_exp} and compute the relative error for these estimators. 

\subsection{AV Traffic Sign Classifier Evaluation}
\label{sec:exp_2}

For this experiment,  we consider the traffic sign classification problem on the German Traffic Sign Benchmark (GTSRB) dataset \cite{stallkamp2011german} which contains about 50,000 images of traffic signs and labels. We evaluate the performance of MicronNet \cite{wong2018micronnet}, one of the state-of-the-art traffic sign classifiers, that achieves 98.7\% accuracy on GTSRB benchmarks. We use traffic sign images sized 32 by 32 pixels.

Our Deep IS framework uses a 4-layer feed-forward ReLU-activated neural network with 32, 16, 8, and 16 neurons in each layer as $g$. We construct a training set with 40,000 training data, which classifies correctly- or incorrectly-classified unperturbed samples with 96.5\% accuracy. We then perturb a fixed correctly-classified input from each class with a Gaussian noise sampled from $N(0, \sigma)$ on each pixel. The parameter $\sigma$ corresponds to the rarity level $\gamma$, where $\sigma = 1/\gamma$. Similar setting can be found in earlier work, e.g. \cite{bai2021over}. We test a rarity level $\gamma=5.0$, which corresponds to $\sigma=0.2$.

\begin{figure*}
    \centering
    \includegraphics[width=\linewidth]{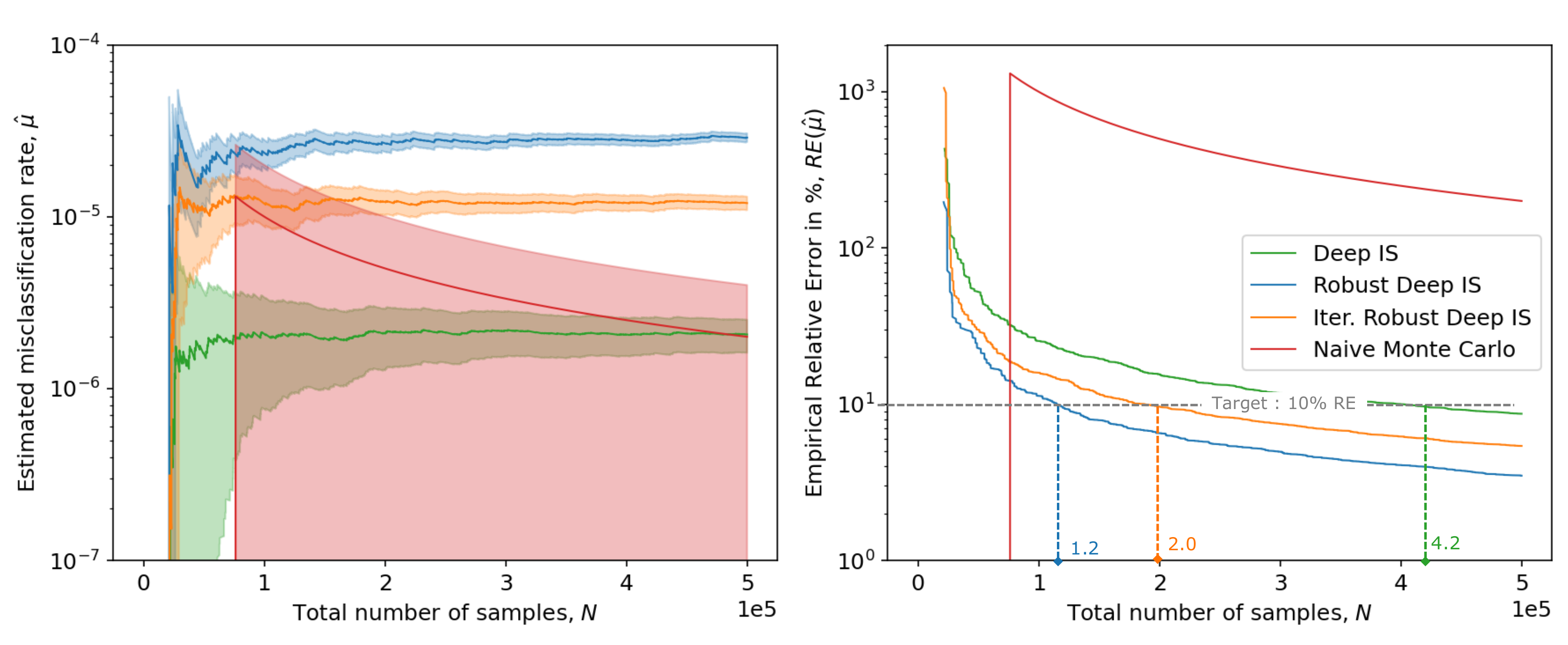}
    \caption{Estimation of AV traffic sign classifier misclassification rate ($\gamma=5.0$)}
    \label{fig:trafficsign_exp}
\end{figure*}

We then run Alg. \ref{alg:deep_is}.  We implement the sequential searching algorithm with MIP formulation and use the first 100 dominating points that we obtain (which takes 72 hours). Due to the high-dimensionality of the input space and the choice of $g$, the number of dominating points in this problem is extremely large (much larger than 100). However, we terminate the process early and only use the first 100 dominating points to construct our proposal distribution $\tilde p$. We also run NMC, Robust Deep IS, and Iter. Robust Deep IS as benchmark. We summarize the estimated misclassification rates and REs in Fig. \ref{fig:trafficsign_exp} and report the number of samples to achieve 10\% RE in Table \ref{tab1}.

\begin{figure}
    \centering
    \includegraphics[width=\linewidth]{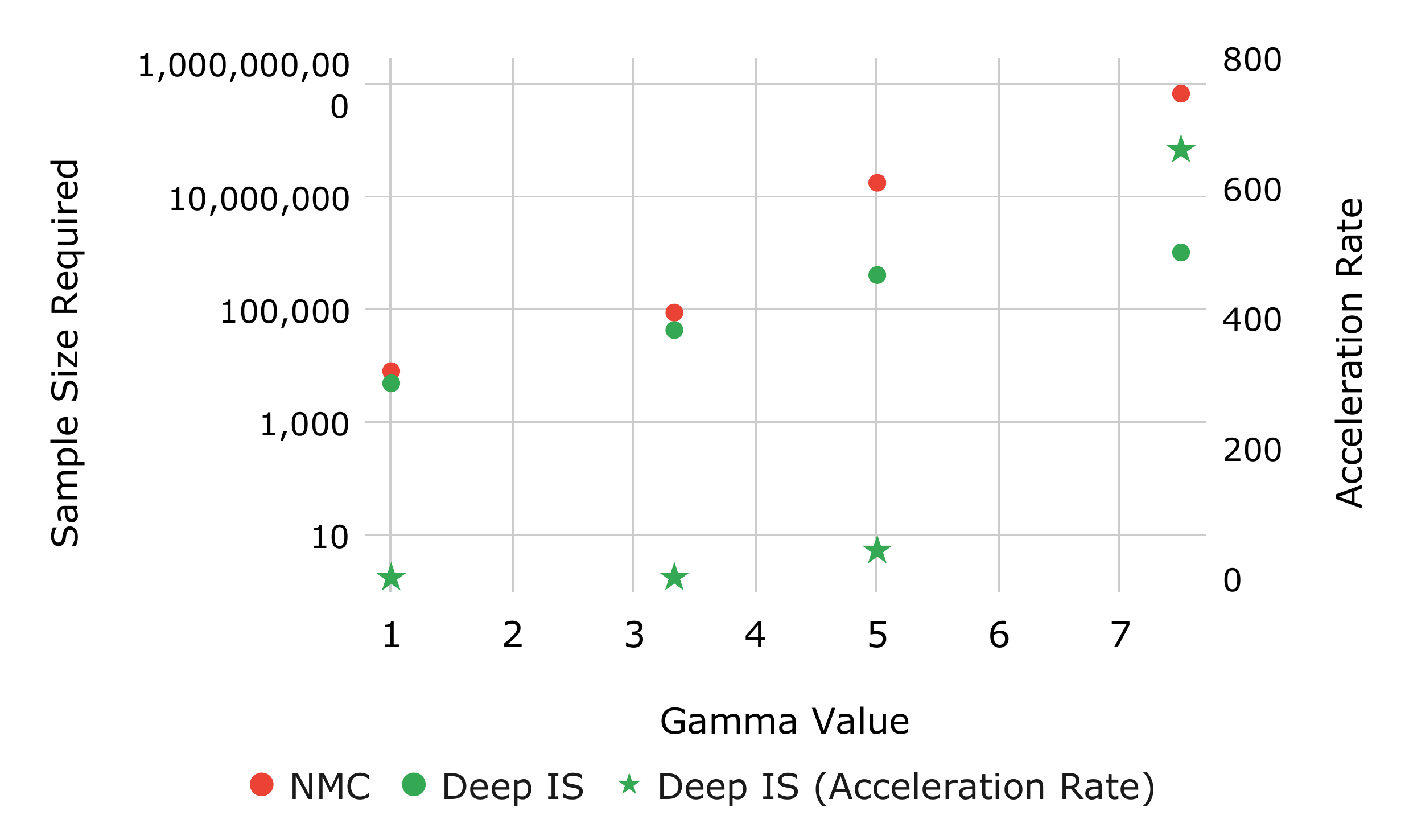}
    \caption{Number of samples required by NMC and Deep IS methods to achieve 10\% RE as rarity parameter $\gamma$ grows}
    \label{fig:num_samples}
\end{figure}

In addition, we also run an ablation study, testing different values of rarity parameter $\gamma = 1.0, 3.33, 5.0, 8.0$, corresponding to $\sigma = 1.0, 0.3, 0.2, 0.125$, respectively. We report the sample size required to achieve 10\% RE by Deep IS and NMC and the corresponding acceleration rate in Fig.  \ref{fig:num_samples}. We note however that NMC sample size for $\gamma=8.0$ is only an approximation, obtained using Eq. \ref{eq:nsample}, due to  the growing computational burden for NMC approach. 

\begin{table*}[t]
\begin{center}
\caption{Sample Size, Degree of Conservativeness, and Acceleration Rate of NMC and Deep IS methods}
\label{tab1}
\begin{tabular}{|lccc|ccc|}
\hline
\multicolumn{1}{|c|}{\multirow{2}{*}{Estimator}} & \multicolumn{3}{c|}{AV Safety Evaluation}                   & \multicolumn{3}{c|}{Traffic Sign Classifier Evaluation ($\gamma=5.0$)}                   \\ \cline{2-7} 
\multicolumn{1}{|c|}{}                           & \multicolumn{1}{c|}{Sample Size} & \multicolumn{1}{c|}{Deg. of Conservativeness} & Acc. Rate & \multicolumn{1}{c|}{Sample Size} & \multicolumn{1}{c|}{Deg. of Conservativeness} & Acc. Rate \\ \hline
NMC                                              & \multicolumn{1}{|c|}{431,210}      & \multicolumn{1}{c|}{1.0 (baseline)} & 1.0 (baseline)     & \multicolumn{1}{c|}{18,181,798} & \multicolumn{1}{c|}{1.0 (baseline)} & 1.0 (baseline)     \\ %\hline
Deep IS                                          & \multicolumn{1}{|c|}{178,832}      & \multicolumn{1}{c|}{1.01} & 2.41                & \multicolumn{1}{c|}{418,983}     & \multicolumn{1}{c|}{1.02} & 43.4               \\ %\hline
Robust Deep IS                                        & \multicolumn{1}{|c|}{32,044}      & \multicolumn{1}{c|}{5.33} & 13.46                & \multicolumn{1}{c|}{118,370}     & \multicolumn{1}{c|}{14.26} & 153.6             \\
Iter. Robust Deep IS                                        & \multicolumn{1}{|c|}{40,231}      & \multicolumn{1}{c|}{2.49} & 10.72                & \multicolumn{1}{c|}{199,372}     & \multicolumn{1}{c|}{9.54} & 91.2             \\ \hline
\end{tabular}
\end{center}
\end{table*}

\section{Discussions}
\label{sec:discussion}

We discuss our findings in terms of estimator's unbiasedness, sampling efficiency, and limitation of the current approach.

\subsection{Estimator's Unbiasedness} 
We first observe from Fig. \ref{fig:av_exp} (left) that Deep-IS estimator (solid green line) and Naive Monte Carlo estimator (solid red line) converge to approximately the same value ($\hat \mu \approx 2 \times 10^{-5}$), showing that Deep IS provides a close estimate of the target risk. Meanwhile the risk upper bound estimators (Robust Deep IS and Iter. Robust Deep IS) settle at $1.1 \times 10^{-4}$  and $6.7 \times 10^{-5}$ respectively, validating the upper bounds they produce. The iterative sampling procedure for the set-learning procedure succeeds in producing a much tighter upper bound (about 50\% closer to the target), though the computation time roughly doubles. This suggests that efforts can be made to produce tighter upper-bound at the cost of extra computation time, which could be helpful if the upper-bound is overly conservative for the task at hand.

This observation repeats in Fig. \ref{fig:trafficsign_exp} (left) when evaluating AV traffic sign classifier. With $\gamma = 5.0$, the misclassification rate is about $2 \times 10^{-6}$, according to Deep IS estimator. Naive Monte Carlo estimator only encounters 1 misclassification case in the whole 100,000 samples during evaluation, resulting in overly high estimation variance. The rarity of the misclassification cases in this example is due to overall smaller noise variance added to the traffic signs, hence the classifier can identify most of the traffic signs correctly using the trained model. Again, the two risk upper-bound estimators provide valid upper bound with respect to the target in these cases. This hints the unbiasedness of Deep IS estimator,  both in the lower-dimensional problem as well as the high-dimensional problem, attaining the lowest degree of conservativeness, compared to the other benchmark.

Table \ref{tab1} highlights the different degrees of conservativeness for all the methods. Deep IS achieves the best performance, with degree of conservativeness about 1, highlighting its close-to-the-target estimates. Meanwhile, the other methods (Robust Deep IS and Iterative Robust Deep IS) fall short, achieving an estimate that is up to 14.26  times higher than the target, highlighting their limitations of only outputting conservative estimates of the true risk, which might not be adequate to make decisions regarding the safety of AV designs.

\subsection{Sampling Efficiency} 
We now focus on the right figure in Fig. \ref{fig:av_exp} and Fig. \ref{fig:trafficsign_exp}, all highlighting how efficient each sampling algorithm is. In  Fig. \ref{fig:av_exp} (right), we observe that Naive Monte Carlo's RE shrinks much slower compared to the other methods, followed by Deep IS, Iter. Robust Deep IS, and Robust Deep IS. While it is clear that Naive Monte Carlo slow shrinkage is due to its sampling efficiency, the case for Deep IS is slightly different. The RE computes the uncertainty of the estimator as a percentage of its estimated value. This means that the larger the estimator is, the larger uncertainty (in magnitude) is allowed. Looking at the \ref{fig:av_exp} (left), we see that the confidence interval of Deep IS estimator is about the same as that of Robust Deep IS and Iter. Robust Deep IS. Thus, the large relative error of Deep IS in \ref{fig:av_exp} (right) is mainly due to its smaller estimated value compared to  the upper bounds that Robust Deep IS and Iter. Robust Deep IS estimate. 

Regardless, suppose now that our goal is to produce an estimate with 20\% maximum RE. Then, Naive Monte Carlo will stop after using 431,210 samples, 178,825 samples for Deep IS, 32,044 samples for Robust Deep IS, and 40,231 for Iter. Robust Deep IS, for the AV experiment. We can determine the sample size  for the traffic sign classifier experiments similarly. Table \ref{tab1} summarizes the sample size. In the table, we also compute the acceleration rate, i.e. the ratio between Naive Monte Carlo (NMC) sample size to the other methods. We immediately found that only Deep IS can obtain relatively unbiased estimate and accelerates the procedure significantly, to up to 43.4 times (for the traffic sign classification with $\gamma  = 5.0$). Meanwhile, the other two variants of Deep IS yields larger acceleration rate, but only capable of obtaining upper bound for the risk. This highlights the tradeoff between the efficiency boost and unbiasedness of Deep IS methods.

\subsection{Deep IS Efficiency vs Rarity Parameter $\gamma$}
We further find that the acceleration rate of Deep IS scales well in the rarity of the target value. Fig. \ref{fig:num_samples} from our ablation study that tests various rarity parameter values finds that for not-so-rare problem, Deep IS acceleration rate (only around 2) can be rather unappealing, considering all the computation resources needed to train the neural network and to solve for the dominating points. However, as the problem grows rarer, the acceleration boost grows very quickly, achieving up to 659 (for $\gamma = 8.0$). This means Deep IS can accelerate the sampling process for evaluation hundreds of times, which can help improve the efficiency of evaluation methods, especially as the AV performance becomes better in the future. We envision that Deep IS efficiency benefit can help the AV community design, evaluate, and deploy safer AVs and reap its full potential for the society at large.

%\subsection*{Achieving higher efficiency boost} 
%We now utilize the upper bound from the Robust Deep IS and Iter. Robust Deep IS as a stopping criterion. Since they provide an upper for the risk, we can safely conclude that the true target is below this upper bound. Thus, we can terminate the procedure when the Robust Deep IS estimate attains the target RE, but outputs Deep IS estimate. We call this Acc. Deep IS. This gives us additional 2-5 times efficiency boost compared to the original Deep IS approach, as summarized in Table \ref{tab2}.  The idea is by realizing that our risk upper-bound estimator has achieved the target RE and still attains much To conclude, our investigation shows evidence of the practical benefits of using Deep-IS for rare-event estimation. It generates valid target for the target probability with low RE and improved efficiency. 

\section{Conclusion and Outlook }
\label{sec:conclusion}

In this paper, we proposed Deep IS, a practical framework to estimate rare-event probabilities for AV evaluation, that is suitable against higher-dimensional uncertainties, often limiting the practicality and efficiency of the state-of-the-art approaches. By using deep neural network approximation, Deep IS achieves hundreds of times efficiency improvement compared to standard sampling approach, when dealing with rare-event problem, while outputs a close-to-target estimate. The latter benefit extends the state-of-the-art certifiable evaluation method for AVs that is only capable of outputting an upper-bound to the target value, limiting its practicality. In the future, we plan to use the upper-bound estimator in parallel to Deep IS estimator to design a more efficient sampling stopping criteria, combining the benefit of the two estimators to further accelerate the efficiency of certifiable evaluation methods for intelligent transportation systems.

%%acknowledgement
\section*{ACKNOWLEDGMENT}
We gratefully acknowledge support from the National Science Foundation under grants CAREER CMMI-1834710, IIS-1849280, IIS-1849304, and CAREER CNS-2047454.

\bibliographystyle{IEEEtran}
\bibliography{bib}

\end{document}